%% file: main_v3.tex
\documentclass[10pt,letterpaper,twocolumn,english]{article}

\usepackage[T1]{fontenc}
\usepackage[latin9]{inputenc}
\usepackage{array}
\usepackage{float}
\usepackage{multirow}
\usepackage{amsmath}
\usepackage{amssymb}
\usepackage{graphicx}

\makeatletter

\pdfpageheight\paperheight
\pdfpagewidth\paperwidth

\providecommand{\tabularnewline}{\\}

\usepackage{iccv}
\usepackage{times}
\usepackage{epsfig}
\usepackage{graphicx}
\usepackage{amsmath}
\usepackage{amssymb}

\usepackage[pagebackref=true,breaklinks=true,letterpaper=true,colorlinks,bookmarks=false]{hyperref}

\iccvfinalcopy 

\def\iccvPaperID{8715} 

\ificcvfinal\fi

\makeatother

\usepackage{babel}
\begin{document}
\title{Few-shot Algorithm Assurance}
\author{Dang Nguyen, Sunil Gupta\\
Applied Artificial Intelligence Institute (A\textsuperscript{2}I\textsuperscript{2}),
Deakin University, Geelong, Australia\\
\textit{\{d.nguyen, sunil.gupta\}@deakin.edu.au}}
\maketitle
\begin{abstract}
\input{abstract_v3.tex}
\end{abstract}
\global\long\def\transp#1{\transpose{#1}}%
 
\global\long\def\transpose#1{#1^{\mathsf{T}}}%
\global\long\def\argmax#1{\underset{#1}{\text{argmax}\ } }%

\section{Introduction\label{sec:Introduction}}

\input{introduction_v3.tex}

\section{Related Background\label{sec:Related-Work}}

\input{relatedwork_v3.tex}

\section{The Proposed Framework\label{sec:Framework}}

\input{framework_v3.tex}

\section{Experiments and Discussions\label{sec:Experiments}}

\input{experiment_v3.tex}

\section{Conclusion\label{sec:Conclusion}}

\input{conclusion_v3.tex}

\bibliographystyle{ieee_fullname}
\bibliography{iccv2023}

\end{document}

%% file: abstract_v3.tex
In image classification tasks, deep learning models are vulnerable
to image distortion. For successful deployment, it is important to
identify distortion levels under which the model is usable i.e. its
accuracy stays above a stipulated threshold. We refer to this problem
as \textit{Model Assurance under Image Distortion}, and formulate
it as a classification task. Given a distortion level, our goal is
to predict if the model's accuracy on the set of distorted images
is greater than a threshold. We propose a novel classifier based on
a Level Set Estimation (LSE) algorithm, which uses the LSE's mean
and variance functions to form the classification rule. We further
extend our method to a \textit{``few sample''} setting where we
can only acquire few real images to perform the model assurance process.
Our idea is to generate extra synthetic images using a novel Conditional
Variational Autoencoder model with two new loss functions. We conduct
extensive experiments to show that our classification method significantly
outperforms strong baselines on five benchmark image datasets.

%% file: introduction_v3.tex
Deep learning models are being increasingly developed and sold by
many AI companies for decision making tasks in diverse domains. They
are often trained with data usually collected under uncontrolled settings
and may have imperfections e.g. noise, bias. The training process
may be imperfect due to various constraints e.g. limited time-budget,
small hardware. Finally, the trained models can be deployed in diverse
conditions e.g. different views, at different scales, different environmental
conditions, and not all of them are appropriately represented in the
training data. Therefore, \textit{at the customer side, it is important
to verify/assure if the deep models work as intended, especially in
scenarios where they are deployed in crucial applications}.

\begin{figure}[th]
\begin{centering}
\includegraphics[scale=0.4]{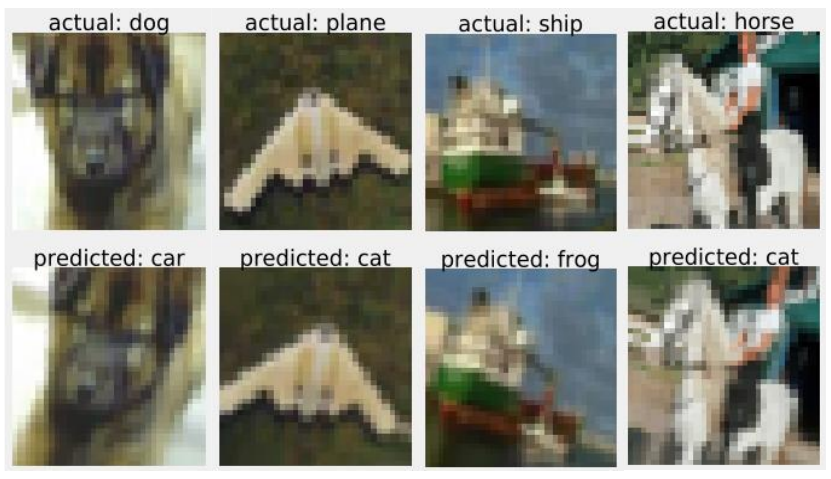}
\par\end{centering}
\caption{\label{fig:Model's-predictions-rotate}Model misclassifies images
rotated by $20^{\circ}$. }
\end{figure}

Assurance can be general or specific to a goal. In this paper, we
want to assure if an image classifier is ``accurate'' under various
image distortions e.g. different views, lighting conditions, etc.
The reason is that image classifiers often perform poorly under image
distortion \cite{li2019blind}. As an example, a ResNet-20 model \cite{he2016deep}
trained on CIFAR-10, achieves 91.37\% accuracy on a validation set.
When this model is deployed in practice, the input images are not
always upright like the original training images, but can be slightly
rotated due to an unstable camera. In this case, the model's performance
can significantly degrade. As shown in Figure \ref{fig:Model's-predictions-rotate},
the model predicts wrong labels for 20-degree rotated images although
these images are easily recognized by humans. The distorted images
are more challenging for the model to classify them correctly, leading
to an accuracy drop in 10\%. The same problem can also happen when
the model was trained with daylight images but deployed in night-time
where the input images are often darker. Thus, \textit{it is important
for end-users to identify the distortion levels under which their
purchased model is usable i.e. its accuracy stays above a stipulated
threshold}.

We refer to the above problem as \textit{Model Assurance under Image
Distortion} (MAID), and form it as a binary classification task. Assume
that we have a pre-trained deep model $T$ and a set of labeled images
${\cal D}$ (we call it \textit{assurance set}). We define \textit{the
search space of distortion levels} ${\cal C}$ as follows: (1) each
dimension of $\mathcal{C}$ is a \textit{type of distortion} (e.g.
rotation, brightness) and (2) each point $c\in\mathcal{C}$ is a\textit{
distortion level} (e.g. \{rotation=10, brightness=1.2\}), which is
used to modify the images in ${\cal D}$ to create a set of distorted
images ${\cal D}'_{c}$. Our goal is to learn a binary classifier
that receives a distortion level $c\in{\cal C}$ as input and predicts
if the model's accuracy on ${\cal D}_{c}'$ is greater than a threshold
$h$. For example, given $h=$ 85\%, $c=$\{rotation=10, brightness=1.2\}
(i.e. it rotates the images $10^{\circ}$ and brightens them by 20\%)
will be classified as ``1'' (``positive'') if $T$ achieves an
accuracy $a\geq85\%$ on ${\cal D}'_{c}$. Otherwise, $c$ will be
classified as ``0'' (``negative'').

To train the classifier, we require a training set. One way to construct
the training set is to randomly sample distortion level $c_{i}$ from
the search space ${\cal C}$, and then compute the accuracy $a_{i}$
of the model $T$ on the set of distorted images ${\cal D}'_{c_{i}}$.
If $a_{i}\geq h$, we assign a label ``1''. Otherwise, we assign
a label ``0''. Finally, we obtain the training set ${\cal R}=\{c_{i},\mathbb{I}_{a_{i}\geq h}\}_{i=1}^{I}$,
where $\mathbb{I}_{a_{i}\geq h}$ is an indicator function and $I$
is the sampling budget. We then use ${\cal R}$ to train a machine
learning classifier e.g. neural network. However, using random distortion
levels to construct the training set ${\cal R}$ is not effective
as often a majority of the distortion levels lead to low accuracy
and thus falling under \textquotedbl negative\textquotedbl{} class.
This leads to an \textit{unbalanced} training set and a \textquotedbl one-sided\textquotedbl{}
\textit{inaccurate} classifier.

To address the above challenge, we propose a novel classifier based
on the \textit{Level Set Estimation} (LSE) algorithm \cite{bryan2005active}.
First, we consider the mapping from a distortion level $c$ to the
model's accuracy on the set of distorted images ${\cal D}'_{c}$ as
a \textit{black-box, expensive function} $f:{\cal C}\rightarrow[0,1]$.
The function $f$ is black-box as we do not know its expression, and
$f$ is expensive as we have to measure the model's accuracy over
all distorted images in ${\cal D}'_{c}$. Second, we approximate $f$
using a Gaussian process (GP) \cite{snoek2012practical} that is a
popular method to model black-box, expensive functions. Third, we
use $f$'s predictive distribution in an acquisition function to search
for distortion levels to update the GP. As we focus on sampling distortion
levels whose predictive function values are close to the threshold,
we have more chance to obtain ``positive'' samples. Finally, we
use the mean and variance returned by the GP to form a classification
rule to predict the label for any point in the search space ${\cal C}$.

We further extend our method to \textit{``few sample''} setting
where only \textit{few original images} are available in ${\cal D}$
(equivalently, we can acquire only \textit{few distorted images}).
Given a distortion level $c$, we typically need lots of distorted
images to accurately estimate the model $T$'s accuracy $a$ on ${\cal D}'_{c}$.
Otherwise, $a$ can be inaccurate, which can be potentially dangerous
in the context of model assurance. For example, the model may achieve
100\% accuracy on 10 distorted images but its true accuracy on a large
set of images at the same distortion level may be lower than the stipulated
threshold. As a result, $c$ may be inaccurately classified as ``positive''
while in truth its label is ``negative''. As collecting a large
number of images is infeasible in some domains e.g. medical or satellite
images,\textit{ model assurance with few images is an important, yet
challenging problem}.

To solve this problem, we try to expand the assurance set ${\cal D}$
with extra synthetic images. In particular, we introduce two new losses
(\textit{distribution} and \textit{prediction losses}) in a Conditional
Variational Autoencoder (CVAE) model \cite{sohn2015learning} to adapt
it to our purpose. The distribution loss tries to match the distribution
of the generated images with that of the training images of the model
$T$. \textit{We note that $T$'s training images are usually inaccessible
(e.g. due to privacy concern)}. The prediction loss tries to generate
images that are recognizable by the model $T$. Using these losses,
our generated images are more diverse and natural, which allows our
LSE-based classifier to be more accurate.

To summarize, we make the following contributions.

(1) We address the problem \textit{Model Assurance under Image Distortion}
(MAID), by proposing a LSE-based classifier to predict if the model
is usable against image distortion.

(2) We further extend our method to a \textit{``few sample''} setting,
by developing a novel generative model to synthesize high-quality
and diverse images.

(3) We extensively evaluate our method on five benchmark image datasets,
and compare it with strong baselines. Our method is significantly
better than other methods.

(4) The significance of our work lies in providing ability to assure
\textit{any} image classifier $T$ under a variety of distortions.
Our method can work irrespective of whether or not $T$ is able to
provide any uncertainty in its decisions.

%% file: relatedwork_v3.tex
\textbf{Image distortion.} Most deep learning models are sensitive
to image distortion, where a small amount of distortion can severely
reduce their performance. Many methods have been proposed to detect
and correct the distortion in the input images \cite{ahn2017image,li2019blind},
which can be categorized into two groups: non-reference and full-reference.
The non-reference methods corrected the distortion without any direct
comparison between the original and distorted images \cite{kang2014convolutional,bosse2016deep}.
Other works developed models that were robust to image distortion,
where most of them fine-tuned the pre-trained models on a pre-defined
set of distorted images \cite{zhou2017classification,dodge2018quality,hossain2019distortion}.
While these methods focused on the space of individual images to improve
the model's robustness, our work focuses on efficiently finding out
the distortions where a model does not meet a required accuracy level.
EXP3BO \cite{gopakumar2018algorithmic} used Bayesian optimization
to verify a model's performance against image distortion. It however
only focused on finding the worst-case performance of the model and
assumed availability of a large assurance set. In contrast, our method
develops a classifier that predicts whether the model's performance
is good or bad at any distortion level in the distortion space and
can work effectively even with a small assurance set.

\textbf{Level set estimation (LSE).} It aims to find the region where
the value of a function is higher or lower than a given threshold
\cite{bryan2005active}. It has many real-world applications, including
environment monitor, quality control, and biotechnology \cite{bryan2005active,bogunovic2016truncated,inatsu2019active}.
LSE was also extended to complex problems e.g. composite functions
\cite{bryan2008actively}, implicit problems \cite{gotovos2013lse},
uncertain inputs \cite{inatsu2020active}, and high-dimensional inputs
\cite{ha2021high}. \textit{However, there is no work applying LSE
to the context of model assurance under image distortion}.

\textbf{Learning with few samples.} Machine learning models often
require many training samples to achieve a good accuracy, a condition
that is rarely met in practice. Few samples learning (FSL) was proposed
to address this problem \cite{wang2020generalizing}. Given a small
set of training images, FSL aims to generate synthetic images to expand
the training set, which helps the model to improve its generalization
and performance. Several common ways to synthesize images include
MixUp \cite{wang2020neural}, CVAE \cite{xu2022generating}, and CGAN
\cite{mirza2014conditional}, which mainly solve supervised learning
problems. \textit{We are the first to tackle the ``few sample''
setting for the model assurance problem}.

%% file: framework_v3.tex
\subsection{Model assurance under image distortion\label{subsec:full-set-maid}}

Let $T$ be a pre-trained deep model (note that $T$'s training set
is not accessible), ${\cal D}=\{x_{i},y_{i}\}_{i=1}^{N}$ be a set
of labeled images (i.e. \textit{assurance set}), and ${\cal S}=\{S_{1},...,S_{d}\}$
be a set of image distortions e.g. rotation, scale, brightness. Each
$S_{i}$ has a value range $[l_{S_{i}},u_{S_{i}}]$, where $l_{S_{i}}$
and $u_{S_{i}}$ are the lower and upper bounds. We define a compact
subset ${\cal C}$ of $\mathbb{R}^{d}$ as a set of all possible values
for image distortion (i.e. ${\cal C}$ is the search space of all
possible distortion levels).

\textbf{Problem statement.} We consider a black-box expensive function
$f:{\cal C}\rightarrow[0,1]$. The function $f(c)$ receives a distortion
level $c$ as input and returns the model's accuracy on the set of
distorted images ${\cal D}_{c}'=\{x_{i}',y_{i}\}_{i=1}^{N}$ as output.
Here, each image $x_{i}'\in{\cal D}_{c}'$ is a distorted version
of an original image $x_{i}\in{\cal D}$. Our goal is to classify
all samples in ${\cal C}$ into ``positive'' and ``negative''
samples based on a user-defined threshold $h\in[0,1]$. More precisely,
given $\forall c\in{\cal C}$, we will assign a ``positive'' label
to $c$ if $f(c)\geq h$. Otherwise, $c$ will be assigned a ``negative''
label. We refer this problem as \textit{Model Assurance under Image
Distortion} (MAID).

As discussed earlier, to address the MAID problem, we can train a
classifier using ${\cal R}=\{c_{i},\mathbb{I}_{f(c_{i})\geq h}\}_{i=1}^{I}$,
where $c_{i}$ is randomly sampled from ${\cal C}$ and $I$ is the
sampling budget. However, ${\cal R}$ is often an \textit{unbalanced}
dataset, where the number of ``negative'' samples is much more than
the number of ``positive'' samples, leading to an \textit{inaccurate}
classifier.

\textbf{LSE-based classifier.} Our idea is that we attempt to obtain
more ``positive'' training samples and develop a classification
rule. First, we initialize a training set ${\cal R}_{t}$ by randomly
sampling $t$ distortion levels $[c_{1},...,c_{t}]$ and computing
their function values $\boldsymbol{f}_{1:t}=[f(c_{1}),...,f(c_{t})]$,
where $t$ is a small number. Second, we use ${\cal R}_{t}$ to learn
a Gaussian process (GP) \cite{snoek2012practical} to approximate
$f$. We assume that $f$ is a smooth function drawn from a GP, i.e.
$f(c)\sim\text{GP}(m(c),k(c,c'))$, where $m(c)$ and $k(c,c')$ are
the mean and covariance functions. We compute the \textit{predictive
distribution} for $f(c)$ at any point $c$ as a Gaussian distribution,
with its mean and variance functions:
\begin{align}
\mu_{t}(c) & =\transp{\boldsymbol{k}}K^{-1}\boldsymbol{f}_{1:t}\label{eq:mean_function}\\
\sigma_{t}^{2}(c) & =k(c,c)-\transp{\boldsymbol{k}}K^{-1}\boldsymbol{k}\label{eq:variance_function}
\end{align}
where $\boldsymbol{k}$ is a vector with its $i$-th element defined
as $k(c_{i},c)$ and $K$ is a matrix of size $t\times t$ with its
$(i,j)$-th element defined as $k(c_{i},c_{j})$. Third, we iteratively
update the training set ${\cal R}_{t}$ by adding the new point $\{c_{t+1},f(c_{t+1})\}$
until the sampling budget $I$ is depleted, and at each iteration
we update the GP. Instead of randomly sampling $c_{t+1}$, we maximize
the acquisition function Straddle proposed in the LSE algorithm \cite{bryan2005active}
to select $c_{t+1}$, which has the following form:
\begin{align}
q(c) & =1.96\times\sigma(c)-|\mu(c)-h|,\label{eq:acq-function}\\
c_{t+1} & =\argmax{c\in{\cal C}}q(c)\label{eq:suggestion}
\end{align}
where $\mu(c)$ and $\sigma(c)$ are the predictive mean and standard
deviation from Equations (\ref{eq:mean_function}) and (\ref{eq:variance_function}).
Our sampling strategy achieves two goals: (1) sampling $c$ where
the model\textquoteright s accuracy is close to the threshold $h$
(i.e. small $|\mu(c)-h|$) and (2) sampling $c$ where the model\textquoteright s
accuracy is highly uncertain (i.e. large $\sigma(c)$). As a result,
we obtain more ``positive'' samples to train our GP. Finally, given
any point $c\in{\cal C}$, as its predictive function value is between
$\mu(c)-2\sigma(c)$ and $\mu(c)+2\sigma(c)$ with 95\% probability,
we form our classification rule as follow:
\begin{equation}
\text{label of }c=\begin{cases}
1, & \text{if }\mu(c)-2\sigma(c)\geq h\\
0, & \text{otherwise}
\end{cases}\label{eq:classification-rule}
\end{equation}

\subsection{Model assurance with few images\label{subsec:few-sample-maid}}

We extend our method to the ``few sample'' setting where only few
images are available in the assurance set. In other words, we acquire
only few distorted images for the model evaluation. Our method has
two steps: (1) we generate synthetic images and (2) we use both original
and synthetic images in our LSE-based classifier.

\subsubsection{Generating synthetic images}

We develop a generative model based on CVAE \cite{sohn2015learning}.
We first review the ``CVAE loss'' used in the standard CVAE. We
then describe our two new losses ``distribution loss'' and ``prediction
loss'' to improve CVAE to generate high-quality and diverse images.

\textbf{CVAE loss.} CVAE aims to generate images similar to its real
input images. Using the assurance set ${\cal D}=\{x_{i},y_{i}\}_{i=1}^{N}$,
CVAE trains encoder and decoder networks to learn a distribution of
latent variable $z\in\mathbb{R}^{l}$. The encoder network maps an
input image $x$ along with its label $y$ to the latent vector $z$
that follows $P(z\mid y)$. The decoder network takes $z$ conditioned
on $y$ to reconstruct $x$. The standard loss to train CVAE is as
follows:
\begin{equation}
{\cal L}_{cvae}={\cal L}_{CE}(x,\tilde{x})+{\cal L}_{KL}(Q(z\mid x,y),P(z\mid y)),\label{eq:cvae-loss}
\end{equation}
where ${\cal L}_{CE}(x,\tilde{x})$ is the cross-entropy loss between
the original image $x$ and the reconstructed image $\tilde{x}$ and
${\cal L}_{KL}(Q(z\mid x,y),P(z\mid y))$ is a Kullback--Leibler
divergence between the approximated posterior distribution of $z$
and the prior distribution of $z$ conditioned on $y$. We choose
a Gaussian distribution for $P(z\mid y)$ i.e. $P(z\mid y)\equiv{\cal N}(0,I)$.

\textbf{Distribution loss.} Although CVAE can generate images, the
diversity of generated images is quite limited as its training data
${\cal D}$ is small. Thus, we encourage CVAE to generate more out-of-distribution
images by adding a new\textit{ distribution loss} term. Our idea is
that we match reconstructed images to the images in both ${\cal D}$
and ${\cal D}_{T}$, where ${\cal D}$ is the training images of CVAE
and ${\cal D}_{T}$ is the training images of the pre-trained deep
model $T$ (typically, $|{\cal D}_{T}|\gg|{\cal D}|$). Here, we cannot
access ${\cal D}_{T}$ since the model owner may not disclose their
training data. Hence, we obtain the statistics of ${\cal D}_{T}$
via the widely-used Batch-Norm (BN) layers in $T$. As a BN layer
normalizes the feature maps during training to eliminate covariate
shifts, it computes the channel-wise means and variances of feature
maps \cite{ioffe2015batch,yin2020dreaming}. As a result, we try to
minimize the distance between the feature statistics of reconstructed
images and original images in ${\cal D}_{T}$. For each batch of reconstructed
images $\{\tilde{x}_{i}\}_{i=1}^{B}$ (where $B$ is the batch size),
we put them into $T$ to obtain their feature maps $\{a_{i}\}_{i=1}^{B}$
corresponding to the first convolution layer, and then compute the
mean $\mu(\{a_{i}\})=\frac{1}{B}\sum_{i=1}^{B}a_{i}$ and the variance
$\sigma^{2}(\{a_{i}\})=\frac{1}{B}\sum_{i=1}^{B}(a_{i}-\mu(\{a_{i}\}))^{2}$.
We then extract the running-mean $\mu_{BN}$ and the running-variance
$\sigma_{BN}^{2}$ stored in the first BN layer of $T$. Finally,
we compute our distribution loss as:
\begin{equation}
{\cal L}_{dist}=||\mu(\{a_{i}\})-\mu_{BN}||_{2}+||\sigma^{2}(\{a_{i}\})-\sigma_{BN}^{2}||_{2},\label{eq:dist-loss}
\end{equation}
where $||\cdot||_{2}$ is $l_{2}$ norm. We use the first BN layer
since we want to match the low-level features of the images in ${\cal D}_{T}$,
which are more task-agnostic. In contrast, higher BN layers capture
high-level features that are more task-specific. 

\textbf{Prediction loss.} Besides the diversity, the quality of generated
images is also important. To this end, we introduce another loss term
-- \textit{prediction loss}, which encourages the generated images
look more natural. Remind that we reconstruct an image $\tilde{x}$
using a latent vector $z$ and a label $y$ via the decoder network.
If $\tilde{x}$ is a random looking image and far from realistic,
the model $T$ cannot recognize $\tilde{x}$. In contrast, if $\tilde{x}$
looks real, $T$ should be able to predict a label $\tilde{y}$ for
$\tilde{x}$ the same as $y$. For example, if the input image $x$
is an image of ``dog'', then $y$ is ``dog'' and $T$ should classify
the reconstructed image $\tilde{x}$ as ``dog''. Based on this observation,
we formulate our prediction loss:
\begin{equation}
{\cal L}_{pred}={\cal L}_{CE}(y,T(\tilde{x})),\label{eq:pred-loss}
\end{equation}
where ${\cal L}_{CE}(y,T(\tilde{x}))$ is the cross-entropy loss between
the input label $y$ and the softmax prediction of the model $T$
for the reconstructed image $\tilde{x}$.

\textbf{Final loss.} We combine Equations (\ref{eq:cvae-loss}), (\ref{eq:dist-loss}),
(\ref{eq:pred-loss}) to derive the final loss to train our generative
model:
\begin{equation}
{\cal L}_{final}={\cal L}_{cvae}+{\cal L}_{dist}+{\cal L}_{pred}\label{eq:final-loss}
\end{equation}

After our generative model is trained, we can generate images via
$G(z,y)$, where $z\sim{\cal N}(0,I)$, $y$ is a label, and $G$
is the trained decoder network.

\textbf{Post-processing.} As not all generated images have good visual
quality, we implement a post-processing step to select the good ones.
Assume that we generate $M$ synthetic images $\hat{{\cal D}}=\{\hat{x}_{1},...,\hat{x}_{M}\}$.
We then use $T$ to predict their class probabilities $O=\{T(\hat{x}_{1}),...,T(\hat{x}_{M})\}$,
and measure the confidence of $T$ on the generated images. The confidence
scores are computed as $E=\{e_{1},...,e_{M}\}$, where $e_{i}=\max(T(\hat{x}_{i}))$.
If the confidence score is high, the synthetic image looks more real
as $T$ confidently predicts the label. Finally, we select the synthetic
images whose the confidence scores are greater than a \textit{confidence
level} $\alpha$.

Our final set of synthetic images is computed as:
\begin{equation}
\hat{{\cal D}}_{final}=\{\hat{x}_{i}\in\hat{{\cal D}}\mid e_{i}>\alpha\}\label{eq:post-process}
\end{equation}

\subsubsection{Training our LSE-based classifier}

We augment the set of synthetic images ${\cal \hat{D}}_{final}$ to
the set of original images ${\cal D}$, and use ${\cal D}\cup{\cal \hat{D}}_{final}$
in our LSE-based classifier (as described earlier). With synthetic
images, our classifier significantly improves its performance in the
``few sample'' setting as shown in our experiments.

%% file: experiment_v3.tex
We conduct comprehensive experiments to show that our method is better
than other methods under both settings: \textit{full set of images}
and \textit{few images}.

\subsection{Experiment settings}

\subsubsection{Datasets}

We use five image datasets MNIST, Fashion, CIFAR-10, CIFAR-100, and
Tiny-ImageNet. They are commonly used to evaluate image classification
methods \cite{he2016deep,rawat2017deep,nguyen2021knowledge}.

\subsubsection{Constructing the test set for the MAID problem}

For each image dataset, we use its training set to train the model
$T$ and use its validation set for the assurance set ${\cal D}$
in the MAID problem. Inspired by \cite{gopakumar2018algorithmic},
we assure the model $T$ against five image distortions as shown in
Table \ref{tab:List-of-transformations}. Note that our method is
applicable to \textit{any} distortion types as long as they can be
defined by a range of values.

\begin{table}[th]
\caption{\label{tab:List-of-transformations}List of distortions along with
their domains.}

\centering{}%
\begin{tabular}{|l|r|l|}
\hline 
\textbf{Distortion} & \textbf{Domain} & \textbf{Description}\tabularnewline
\hline 
\hline 
Rotation & $[0,90]$ & Rotate $0^{\circ}$ - $90^{\circ}$\tabularnewline
\hline 
Scale & $[0.7,1.3]$ & Zoom in/out 0-30\%\tabularnewline
\hline 
Translation-X & $[-0.2,0.2]$ & Shift left/right 0-20\%\tabularnewline
\hline 
Translation-Y & $[-0.2,0.2]$ & Shift up/down 0-20\%\tabularnewline
\hline 
Brightness & $[0.7,1.3]$ & Darken/brighten 0-30\%\tabularnewline
\hline 
\end{tabular}
\end{table}

To evaluate the performance of each assurance method, we need to construct
the test set. Inspired by LSE methods \cite{gotovos2013lse,ha2021high},
we create a grid of data points in the search space (here, \textit{each
data point is a distortion level}). For each dimension, we use five
points, resulting in 3,125 grid points in total. For each point $c$,
we use it to modify the images in ${\cal D}$, and compute $T$'s
accuracy on the set of distorted images ${\cal D}_{c}'$. If the accuracy
is greater than a threshold, $c$ is assigned a ``positive'' label.
Otherwise, $c$ has a ``negative'' label. At the end, there are
3,125 test points along with their labels. We report the numbers of
``positive'' and ``negative'' test points for each image dataset
in Appendix 1.

\subsubsection{Evaluation metric}

We use each assurance method to predict labels for 3,125 test points
and compare them with the true labels. Since the number of ``positive''
test points is much smaller than the number of ``negative'' test
points, we use the F1-score (i.e. the harmonic mean of precision and
recall) to evaluate the prediction. The higher the F1-score, the better
in prediction.

\subsubsection{Training the image classifier $T$}

As most AI companies offer general AI-based services, it is reasonable
to assume that the model developers do not know about the deployment
conditions at the customer side. Thus, we use pre-trained models from
\cite{nguyen2022fskd} for $T$, which achieve similar accuracy on
the original validation set as those reported in \cite{tian2020contrastive,wang2020neural,bhat2021distill}.
As pointed out by \cite{kang2014convolutional,hossain2019distortion,li2019blind},
we expect that these models will reduce their performance when evaluated
on \textit{unseen distorted images} (evidenced by the large number
of ``negative'' test points). Note that our method can verify/assure
\textit{any} image classifiers. 

\subsubsection{Baselines}

For the MAID problem (see Section \ref{subsec:full-set-maid}), we
compare our LSE-based classifier (called \textbf{LSE-C}) with five
popular machine learning classifiers, including \textit{decision tree}
(DT), \textit{random forest} (RF), \textit{logistic regression} (LR),
\textit{support vector machine} (SVM), and \textit{neural network}
(NN). Since we are dealing with imbalanced classification, we modify
the loss functions to allocate high/low weights for minority/majority
samples using the inverse class frequency. This approach called \textquotedblleft \textit{cost-sensitive
learning}\textquotedblright{} \cite{thai2010cost}. We also compare
with other well-known techniques to train a classifier with imbalanced
data, including 2 under-sampling methods (\textit{RandomUnder} and
\textit{NearMiss}), 3 over-sampling methods (\textit{RandomOver},
\textit{SMOTE}, and \textit{AdaSyn}), and 2 generative methods (\textit{GAN}
and \textit{VAE}). Their details are in \cite{leevy2018survey,sampath2021survey}.

For MAID with few images (see Section \ref{subsec:few-sample-maid}),
we compare our ``few sample'' version (called \textbf{FS-LSE-C})
with 3 baselines \textit{CVAE-LSE-C}, \textit{CGAN-LSE-C}, and \textit{MixUp-LSE-C}.
Similar to our method, these baselines first use CVAE \cite{sohn2015learning},
CGAN \cite{mirza2014conditional}, or MixUp \cite{zhang2018mixup}
to generate synthetic images, then apply LSE-C to both original and
synthetic images. In this experiment, we want to show that our generative
model is better than the other techniques for image generation, and
helps towards better classification.

For a fair comparison, we use the same assurance set ${\cal D}$ and
sampling budget ($I=400$) for all methods. In the ``few image''
setting, we generate the same number of synthetic images for all methods.
We use the same network architecture and hyper-parameters for CVAE-LSE-C
and our FS-LSE-C. The difference is that CVAE-LSE-C uses the standard
CVAE loss in Equation (\ref{eq:cvae-loss}) while FS-LSE-C uses our
proposed loss in Equation (\ref{eq:final-loss}). We use the confidence
level $\alpha=0.8$ to select qualified generated images across all
experiments. We repeat each method three times with random seeds,
and report its averaged F1-score. We do not report the standard deviations
since they are small ($<0.02$).

\subsection{Results on MNIST and Fashion}

The pre-trained image classifiers $T$ achieve 99.31\% and 92.67\%
accuracy on MNIST and Fashion (the accuracy is measured on the original
validation set). We set the threshold $h=95\%$ for MNIST and $h=85\%$
for Fashion. Both datasets have 10 classes, and each class has 6K
training images, and 1K testing images.

\subsubsection{Results with the ``full set of images'' setting}

We evaluate our method when the assurance set ${\cal D}$ contains
the full set of images (10K images). Table \ref{tab:Results-MNIST-Fashion-full}
shows that our method LSE-C is much better than other methods, where
its improvements are around 7\% (MNIST) and 30\% (Fashion) over the
best baseline. Note that as each imbalance handling technique is combined
with 5 classifiers, we only report the best combination.

\begin{table}[th]
\caption{\label{tab:Results-MNIST-Fashion-full}F1-scores on MNIST and Fashion.
``$N$'' shows the number of original images used by each method.}

\centering{}%
\begin{tabular}{|l|l|r|r|}
\hline 
\textbf{Dataset} & \textbf{Method} & \textbf{$N$} & \textbf{F1-score}\tabularnewline
\hline 
\multirow{9}{*}{MNIST} & Cost-Sensitive & 10,000 & 0.6727\tabularnewline
\cline{2-4} \cline{3-4} \cline{4-4} 
 & RU & 10,000 & 0.2118\tabularnewline
\cline{2-4} \cline{3-4} \cline{4-4} 
 & NearMiss & 10,000 & 0.2789\tabularnewline
\cline{2-4} \cline{3-4} \cline{4-4} 
 & RO & 10,000 & 0.6657\tabularnewline
\cline{2-4} \cline{3-4} \cline{4-4} 
 & SMOTE & 10,000 & 0.6253\tabularnewline
\cline{2-4} \cline{3-4} \cline{4-4} 
 & AdaSyn & 10,000 & 0.6322\tabularnewline
\cline{2-4} \cline{3-4} \cline{4-4} 
 & GAN  & 10,000 & 0.5687\tabularnewline
\cline{2-4} \cline{3-4} \cline{4-4} 
 & VAE & 10,000 & 0.5952\tabularnewline
\cline{2-4} \cline{3-4} \cline{4-4} 
 & \textbf{LSE-C} (Ours) & 10,000 & \textbf{0.7444}\tabularnewline
\hline 
\multicolumn{4}{c}{\vspace{-0.3cm}
}\tabularnewline
\hline 
\multirow{9}{*}{Fashion} & Cost-Sensitive & 10,000 & 0.3481\tabularnewline
\cline{2-4} \cline{3-4} \cline{4-4} 
 & RU & 10,000 & 0.0078\tabularnewline
\cline{2-4} \cline{3-4} \cline{4-4} 
 & NearMiss & 10,000 & 0.0078\tabularnewline
\cline{2-4} \cline{3-4} \cline{4-4} 
 & RO & 10,000 & 0.3538\tabularnewline
\cline{2-4} \cline{3-4} \cline{4-4} 
 & SMOTE & 10,000 & 0.3538\tabularnewline
\cline{2-4} \cline{3-4} \cline{4-4} 
 & AdaSyn & 10,000 & 0.3538\tabularnewline
\cline{2-4} \cline{3-4} \cline{4-4} 
 & GAN  & 10,000 & 0.2169\tabularnewline
\cline{2-4} \cline{3-4} \cline{4-4} 
 & VAE & 10,000 & 0.3481\tabularnewline
\cline{2-4} \cline{3-4} \cline{4-4} 
 & \textbf{LSE-C} (Ours) & 10,000 & \textbf{0.6558}\tabularnewline
\hline 
\end{tabular}
\end{table}

\begin{table}[th]
\caption{\label{tab:Results-MNIST-Fashion-few}F1-scores on MNIST and Fashion.
``$N$'' and ``$M$'' show the number of original and synthetic
images used by each method.}

\centering{}%
\begin{tabular}{|l|l|r|r|r|}
\hline 
\textbf{Dataset} & \textbf{Method} & \textbf{$N$} & $M$ & \textbf{F1-score}\tabularnewline
\hline 
\hline 
\multirow{6}{*}{MNIST} & AdaSyn & 50 & - & 0.5771\tabularnewline
\cline{2-5} \cline{3-5} \cline{4-5} \cline{5-5} 
 & \textbf{LSE-C} (Ours) & 50 & - & 0.5321\tabularnewline
\cline{2-5} \cline{3-5} \cline{4-5} \cline{5-5} 
 & CVAE-LSE-C & 50 & 500 & 0.0417\tabularnewline
\cline{2-5} \cline{3-5} \cline{4-5} \cline{5-5} 
 & CGAN-LSE-C & 50 & 500 & 0.0308\tabularnewline
\cline{2-5} \cline{3-5} \cline{4-5} \cline{5-5} 
 & MixUp-LSE-C & 50 & 500 & 0.0923\tabularnewline
\cline{2-5} \cline{3-5} \cline{4-5} \cline{5-5} 
 & \textbf{FS-LSE-C }(Ours) & 50 & 500 & \textbf{0.5951}\tabularnewline
\hline 
\multicolumn{3}{c}{\vspace{-0.3cm}
} & \tabularnewline
\hline 
\multirow{6}{*}{Fashion} & SMOTE & 50 & - & 0.3446\tabularnewline
\cline{2-5} \cline{3-5} \cline{4-5} \cline{5-5} 
 & \textbf{LSE-C} (Ours) & 50 & - & 0.4444\tabularnewline
\cline{2-5} \cline{3-5} \cline{4-5} \cline{5-5} 
 & CVAE-LSE-C & 50 & 500 & 0.1481\tabularnewline
\cline{2-5} \cline{3-5} \cline{4-5} \cline{5-5} 
 & CGAN-LSE-C & 50 & 500 & 0.0833\tabularnewline
\cline{2-5} \cline{3-5} \cline{4-5} \cline{5-5} 
 & MixUp-LSE-C & 50 & 500 & 0.3846\tabularnewline
\cline{2-5} \cline{3-5} \cline{4-5} \cline{5-5} 
 & \textbf{FS-LSE-C }(Ours) & 50 & 500 & \textbf{0.5808}\tabularnewline
\hline 
\end{tabular}
\end{table}

\subsubsection{Results with the ``few image'' setting}

We evaluate our method when the assurance set ${\cal D}$ contains
only few images (\textit{5 images/class}). We train our generative
model with feed-forward neural networks for encoder and decoder, latent
dim = 2, batch size = 16, and \#epochs = 600. 

As shown in Table \ref{tab:Results-MNIST-Fashion-few}, all methods
significantly drop their performance when run with few images. With
synthetic images, our ``few sample'' version FS-LSE-C greatly improves
its F1-scores over LSE-C and other methods. Here, we only report F1-scores
of AdaSyn and SMOTE (the best imbalance handling methods). Other results
are in Appendix 2.

Other generative-based methods do not show any benefit, where their
performance is much lower than ours. This is because their synthetic
images do not have enough diversity as CVAE and CGAN easily over-fit
and memorize few samples due to their limited training data (only
5 images/class).

\subsection{Results on CIFAR-10 and CIFAR-100}

The pre-trained image classifiers $T$ achieve 91.37\% accuracy on
CIFAR-10 and 69.08\% accuracy on CIFAR-100. We assure $T$ with the
thresholds $h=85\%$ (CIFAR-10) and $h=65\%$ (CIFAR-100). CIFAR-10
contains RGB images with 10 classes, and each class has 5K training
and 1K testing images. CIFAR-100 has 100 classes, and each of them
has 500 training and 100 testing images.

\subsubsection{Results with the ``full set of images'' setting}

From Table \ref{tab:Results-CIFAR-full}, our method LSE-C is the
best method, where its improvements are notable. Cost-Sen is generally
better than other methods. Over-sampling and generative methods always
outperform under-sampling methods.

Note that instead of using random points, we can use the points suggested
by Equation (\ref{eq:suggestion}) to train a machine learning classifier.
We compare with this approach in Appendix 3.

\begin{table}[th]
\caption{\label{tab:Results-CIFAR-full}F1-scores on CIFAR in the ``full set
of images'' setting.}

\centering{}%
\begin{tabular}{|l|l|r|r|}
\hline 
\textbf{Dataset} & \textbf{Method} & \textbf{$N$} & \textbf{F1-score}\tabularnewline
\hline 
\hline 
\multirow{9}{*}{CIFAR-10} & Cost-Sensitive & 10,000 & 0.8549\tabularnewline
\cline{2-4} \cline{3-4} \cline{4-4} 
 & RU & 10,000 & 0.7524\tabularnewline
\cline{2-4} \cline{3-4} \cline{4-4} 
 & NearMiss & 10,000 & 0.7923\tabularnewline
\cline{2-4} \cline{3-4} \cline{4-4} 
 & RO & 10,000 & 0.8489\tabularnewline
\cline{2-4} \cline{3-4} \cline{4-4} 
 & SMOTE & 10,000 & 0.8570\tabularnewline
\cline{2-4} \cline{3-4} \cline{4-4} 
 & AdaSyn & 10,000 & 0.8563\tabularnewline
\cline{2-4} \cline{3-4} \cline{4-4} 
 & GAN & 10,000 & 0.8551\tabularnewline
\cline{2-4} \cline{3-4} \cline{4-4} 
 & VAE & 10,000 & 0.8567\tabularnewline
\cline{2-4} \cline{3-4} \cline{4-4} 
 & \textbf{LSE-C} (Ours) & 10,000 & \textbf{0.9098}\tabularnewline
\hline 
\multicolumn{4}{c}{\vspace{-0.3cm}
}\tabularnewline
\hline 
\multirow{9}{*}{CIFAR-100} & Cost-Sensitive & 10,000 & 0.7115\tabularnewline
\cline{2-4} \cline{3-4} \cline{4-4} 
 & RU & 10,000 & 0.2016\tabularnewline
\cline{2-4} \cline{3-4} \cline{4-4} 
 & NearMiss & 10,000 & 0.2246\tabularnewline
\cline{2-4} \cline{3-4} \cline{4-4} 
 & RO & 10,000 & 0.7115\tabularnewline
\cline{2-4} \cline{3-4} \cline{4-4} 
 & SMOTE & 10,000 & 0.5540\tabularnewline
\cline{2-4} \cline{3-4} \cline{4-4} 
 & AdaSyn & 10,000 & 0.5538\tabularnewline
\cline{2-4} \cline{3-4} \cline{4-4} 
 & GAN & 10,000 & 0.6207\tabularnewline
\cline{2-4} \cline{3-4} \cline{4-4} 
 & VAE & 10,000 & 0.5751\tabularnewline
\cline{2-4} \cline{3-4} \cline{4-4} 
 & \textbf{LSE-C} (Ours) & 10,000 & \textbf{0.8000}\tabularnewline
\hline 
\end{tabular}
\end{table}

\begin{table}[th]
\caption{\label{tab:Results-CIFAR-few}F1-scores on CIFAR in the ``few image''
setting.}

\centering{}%
\begin{tabular}{|l|l|r|r|r|}
\hline 
\textbf{Dataset} & \textbf{Method} & \textbf{$N$} & $M$ & \textbf{F1-score}\tabularnewline
\hline 
\hline 
\multirow{6}{*}{CIFAR-10} & AdaSyn & 50 & - & 0.7901\tabularnewline
\cline{2-5} \cline{3-5} \cline{4-5} \cline{5-5} 
 & \textbf{LSE-C} (Ours) & 50 & - & 0.8128\tabularnewline
\cline{2-5} \cline{3-5} \cline{4-5} \cline{5-5} 
 & CVAE-LSE-C & 50 & 500 & 0.0090\tabularnewline
\cline{2-5} \cline{3-5} \cline{4-5} \cline{5-5} 
 & CGAN-LSE-C & 50 & 500 & 0.0089\tabularnewline
\cline{2-5} \cline{3-5} \cline{4-5} \cline{5-5} 
 & MixUp-LSE-C & 50 & 500 & \textbf{0.8697}\tabularnewline
\cline{2-5} \cline{3-5} \cline{4-5} \cline{5-5} 
 & \textbf{FS-LSE-C }(Ours) & 50 & 500 & 0.8687\tabularnewline
\hline 
\multicolumn{3}{c}{\vspace{-0.3cm}
} & \tabularnewline
\hline 
\multirow{6}{*}{CIFAR-100} & RO & 500 & - & 0.2788\tabularnewline
\cline{2-5} \cline{3-5} \cline{4-5} \cline{5-5} 
 & \textbf{LSE-C} (Ours) & 500 & - & 0.3010\tabularnewline
\cline{2-5} \cline{3-5} \cline{4-5} \cline{5-5} 
 & CVAE-LSE-C & 500 & 500 & 0.0000\tabularnewline
\cline{2-5} \cline{3-5} \cline{4-5} \cline{5-5} 
 & CGAN-LSE-C & 500 & 500 & 0.0000\tabularnewline
\cline{2-5} \cline{3-5} \cline{4-5} \cline{5-5} 
 & MixUp-LSE-C & 500 & 500 & 0.4094\tabularnewline
\cline{2-5} \cline{3-5} \cline{4-5} \cline{5-5} 
 & \textbf{FS-LSE-C }(Ours) & 500 & 500 & \textbf{0.5787}\tabularnewline
\hline 
\end{tabular}
\end{table}

\subsubsection{Results with the ``few image'' setting}

We train our CVAE with convolutional neural networks for both encoder
and decoder, with latent dimension = 2, batch size = 64, and \#epochs
= 600.

From Table \ref{tab:Results-CIFAR-few}, our FS-LSE-C is much better
than other generative methods. It is comparable with MixUp-LSE-C on
CIFAR-10 but much better on CIFAR-100. As expected, our LSE-C drops
its F1-score when run with few images, but it still outperforms over-sampling
methods \textasciitilde 2-3\%.

\subsection{Results on Tiny-ImageNet}

For the ``full set of images'' setting (Table \ref{tab:Results-Tiny-IN-full}),
our LSE-C achieves 0.9113 F1-score, which is much better than the
best baseline RO (0.7453). For the ``few image'' setting, our FS-LSE-C
becomes the best, where its F1-score is 0.8287 vs. MixUp-LSE-C (0.6938).
Details are in Appendix 4.

\begin{table}[H]
\caption{\label{tab:Results-Tiny-IN-full}F1-scores on Tiny-IN in the ``full
set of images'' setting.}

\centering{}%
\begin{tabular}{|l|l|r|r|}
\hline 
\textbf{Dataset} & \textbf{Method} & \textbf{$N$} & \textbf{F1-score}\tabularnewline
\hline 
\hline 
\multirow{9}{*}{Tiny-IN} & Cost-Sensitive & 10,000 & 0.7053\tabularnewline
\cline{2-4} \cline{3-4} \cline{4-4} 
 & RU & 10,000 & 0.4207\tabularnewline
\cline{2-4} \cline{3-4} \cline{4-4} 
 & NearMiss & 10,000 & 0.5014\tabularnewline
\cline{2-4} \cline{3-4} \cline{4-4} 
 & RO & 10,000 & 0.7453\tabularnewline
\cline{2-4} \cline{3-4} \cline{4-4} 
 & SMOTE & 10,000 & 0.7267\tabularnewline
\cline{2-4} \cline{3-4} \cline{4-4} 
 & AdaSyn & 10,000 & 0.7385\tabularnewline
\cline{2-4} \cline{3-4} \cline{4-4} 
 & GAN & 10,000 & 0.6687\tabularnewline
\cline{2-4} \cline{3-4} \cline{4-4} 
 & VAE & 10,000 & 0.6635\tabularnewline
\cline{2-4} \cline{3-4} \cline{4-4} 
 & \textbf{LSE-C} (Ours) & 10,000 & \textbf{0.9113}\tabularnewline
\hline 
\end{tabular}
\end{table}

The results in Tables \ref{tab:Results-MNIST-Fashion-full}-\ref{tab:Results-Tiny-IN-full}
suggest that our methods are effective under both settings and always
better than existing methods. With our novel generative model to synthesize
images, our methods solve the model assurance problem more efficient
and realistic even with only few images.

\subsection{Ablation study}

We conduct further experiments on CIFAR-10 to analyze our method FS-LSE-C
under the ``few image'' setting.

\begin{table*}[t]
\caption{\label{tab:Effectiveness-components}Effectiveness of different components
in our method FS-LSE-C.}

\centering{}%
\begin{tabular}{|l|c|c|c|c|c|c|c|c|}
\hline 
 & CVAE-LSE-C & \multicolumn{7}{c|}{FS-LSE-C (Ours)}\tabularnewline
\hline 
\hline 
Distribution loss &  & $\checkmark$ &  &  & $\checkmark$ & $\checkmark$ &  & $\checkmark$\tabularnewline
\hline 
Prediction loss &  &  & $\checkmark$ &  & $\checkmark$ &  & $\checkmark$ & $\checkmark$\tabularnewline
\hline 
Post-processing &  &  &  & $\checkmark$ &  & $\checkmark$ & $\checkmark$ & $\checkmark$\tabularnewline
\hline 
\textbf{F1-score} & 0.0090 & 0.1079 & 0.0090 & 0.7546 & 0.4652 & 0.8223 & 0.8132 & \textbf{0.8687}\tabularnewline
\hline 
\end{tabular}
\end{table*}

\begin{figure*}
\begin{centering}
\includegraphics[scale=0.55]{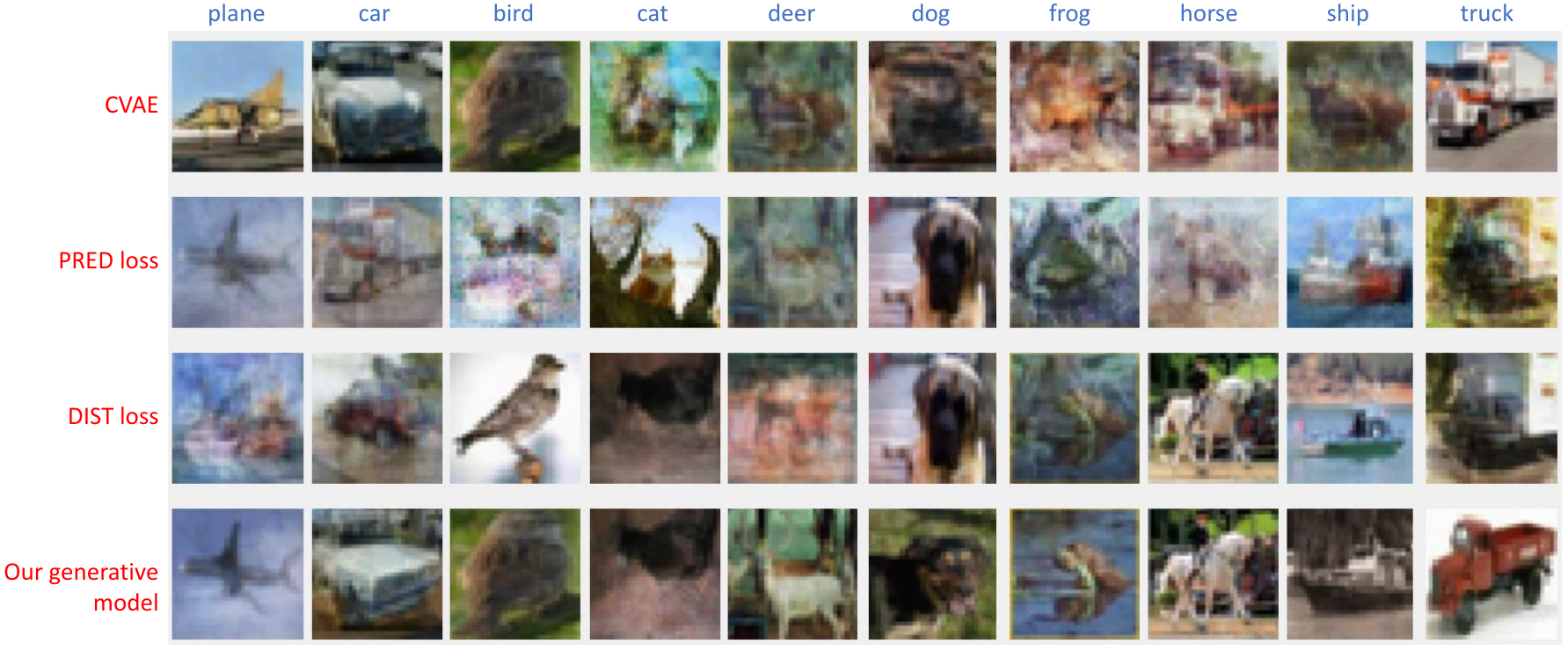}
\par\end{centering}
\caption{\label{fig:Synthetic-images}Synthetic images generated by CVAE and
our generative model with different losses. ``PRED loss'' and ``DIST
loss'' mean our model only uses the prediction loss (Eq. (\ref{eq:pred-loss}))
or the distribution loss (Eq. (\ref{eq:dist-loss})). The text on
the top indicates the labels of generated images.}
\end{figure*}

\subsubsection{Different types of components}

Compared to the standard CVAE, our generative model has three new
components: \textit{distribution loss}, \textit{prediction loss},
and \textit{post-processing step}. Table \ref{tab:Effectiveness-components}
reports the F1-score of each component. While CVAE-LSE-C using CVAE
alone only achieves 0.0090, each individual component in our method
(except the prediction loss) shows a significant improvement. The
post-processing becomes the most successful component, achieving up
to 0.7546. By combining the post-processing with the prediction loss
or the distribution loss, our method further improves its F1-score
up to 0.8132 and 0.8223 respectively. Finally, when using all three
components together, we achieve the best result at 0.8687.

Figure \ref{fig:Synthetic-images} shows synthetic images generated
by CVAE and our generative model with different losses. The CVAE images
are often unclear due to its limited training data. With our prediction
and distribution losses, the synthetic images are slightly better.
Our final generative model generates the best-quality images, where
the subjects (e.g. ``dog'', ``cat'') are clearly recognized and
visualized.

\textbf{FID scores.} We compute the Frechet inception distance (FID)
scores \cite{heusel2017gans} to measure the quality of the generated
images (lower is better): CVAE (6.99), PRED loss (7.56), DIST loss
(5.57), and our\textbf{ }generative model (4.36). 

The ablation study suggests that each component in our generative
model is useful, which greatly improves the classification performance
compared to the standard CVAE. By combining all three components,
our generative model generates high-quality and diverse synthetic
images.

\subsubsection{Hyper-parameter analysis}

There is one important hyper-parameter in our framework, which is
the confidence level $\alpha\in[0,1]$ to select good generated images
in the post-processing step. We investigate how the different values
of $\alpha$ affect our F1-score.

\begin{figure}[H]
\begin{centering}
\includegraphics[scale=0.25]{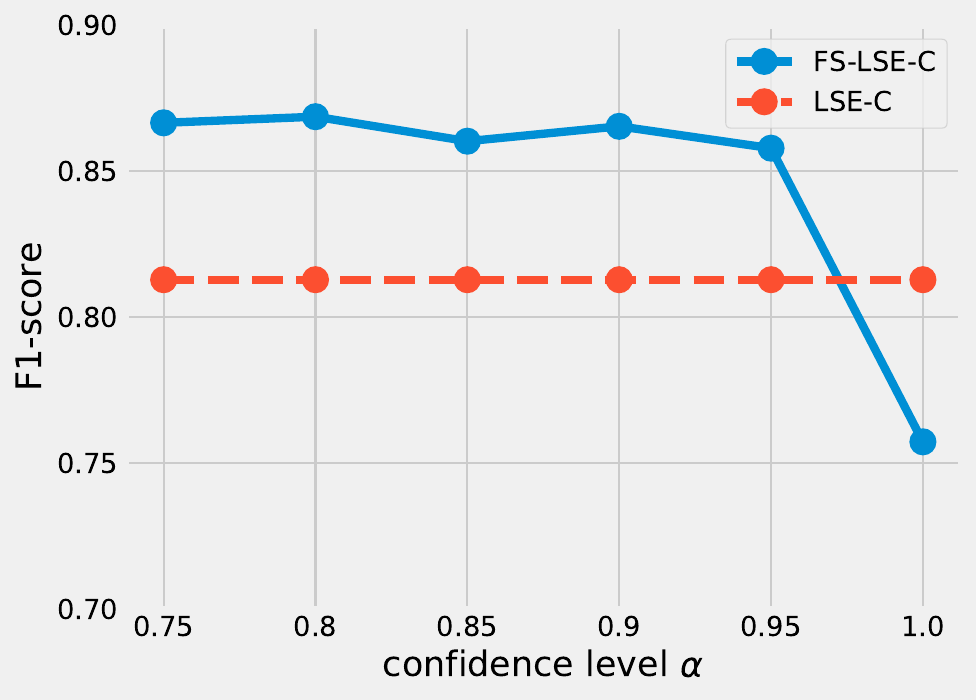}
\par\end{centering}
\caption{\textbf{\label{fig:F1-score-alpha}}F1-score vs. confidence level
$\alpha$ on CIFAR-10.}
\end{figure}

From Figure \ref{fig:F1-score-alpha}, our ``few sample'' version
FS-LSE-C is always better than LSE-C with $\alpha\in[0.75,0.95]$.
More importantly, it is stable with a large range of $\alpha$ values,
where its F1-score is just slightly changed. When $\alpha$ is small
(i.e. $\alpha<0.75$), it may drop its F1-score as most of generated
images are not confidently recognized. When $\alpha$ is too large
(i.e. $\alpha>0.95$), it also reduces its F1-score since very few
synthetic images are generated under this strict constraint.

\subsubsection{Numbers of original and synthetic images}

We investigate the effect of the number of original and generated
images on our method's performance in Appendix 5.

In summary, both methods LSE-C and FS-LSE-C are improved with more
original images. FS-LSE-C is improved with more synthetic images and
is always better than LSE-C (our version does not use synthetic images).

\subsubsection{Visualization}

\begin{figure}[H]
\begin{centering}
\includegraphics[scale=0.7]{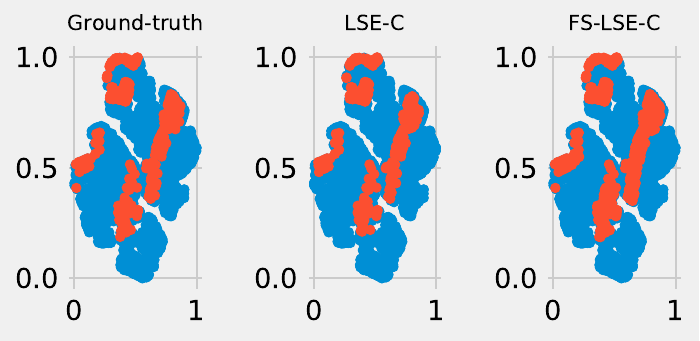}
\par\end{centering}
\caption{\label{fig:Visualization-of-assured-region}``Positive'' and ``negative''
points classified by our methods vs. the ground-truth on CIFAR-10.
Red dots denote ``positive'' points while blue dots denote ``negative''
points.}
\end{figure}

We visualize the ``positive'' and ``negative'' points classified
by our methods using t-SNE \cite{van2008visualizing}, and compare
them with the ground-truth labels in Figure \ref{fig:Visualization-of-assured-region}.
In summary, our methods correctly predict the labels for the test
points.

%% file: conclusion_v3.tex
We address the problem \textit{Model Assurance under Image Distortion}
(MAID) i.e. we seek to find all distortion levels where the accuracy
of a pre-trained deep model exceeds a threshold. By doing this, we
gain knowledge of the model's strengths and weaknesses before deploying
it in real-world. Since MAID can be formed as a classification task,
we proposed a novel classifier based on a LSE algorithm. Our idea
is to approximate the expensive, black-box function mapping from a
distortion level to the model's accuracy on distorted images using
GP. We then leverage the GP's mean and variance to form our classification
rule.

We further extend our method to a ``few sample'' setting where we
can acquire only few real images to perform the model assurance process.
We develop a CVAE-based generative model, with two new loss functions
and a post-processing step. These components allow our generative
model to produce a high-quality and diverse synthetic images, which
significantly improves the performance of our LSE-based classifier.
Our method does not require many labeled samples, thus it can be directly
applied to domains where labeled images are difficult to collect e.g.
medical or satellite images. We demonstrate the benefits of our methods
on five image datasets, where they greatly outperform other methods
including machine learning classifiers with imbalance handling capability
and generative methods.

%% file: main_v3.bbl
\begin{thebibliography}{10}\itemsep=-1pt

\bibitem{ahn2017image}
Namhyuk Ahn, Byungkon Kang, and Kyung-Ah Sohn.
\newblock Image distortion detection using convolutional neural network.
\newblock In {\em IEEE Asian Conference on Pattern Recognition (ACPR)}, pages
  220--225, 2017.

\bibitem{bhat2021distill}
Prashant Bhat, Elahe Arani, and Bahram Zonooz.
\newblock Distill on the go: Online knowledge distillation in self-supervised
  learning.
\newblock In {\em CVPR}, pages 2678--2687, 2021.

\bibitem{bogunovic2016truncated}
Ilija Bogunovic, Jonathan Scarlett, Andreas Krause, and Volkan Cevher.
\newblock Truncated variance reduction: A unified approach to bayesian
  optimization and level-set estimation.
\newblock {\em NIPS}, 29, 2016.

\bibitem{bosse2016deep}
Sebastian Bosse, Dominique Maniry, Thomas Wiegand, and Wojciech Samek.
\newblock A deep neural network for image quality assessment.
\newblock In {\em IEEE International Conference on Image Processing (ICIP)},
  pages 3773--3777, 2016.

\bibitem{bryan2005active}
Brent Bryan, Robert Nichol, Christopher Genovese, Jeff Schneider, Christopher
  Miller, and Larry Wasserman.
\newblock Active learning for identifying function threshold boundaries.
\newblock {\em NIPS}, 18, 2005.

\bibitem{bryan2008actively}
Brent Bryan and Jeff Schneider.
\newblock Actively learning level-sets of composite functions.
\newblock In {\em ICML}, pages 80--87, 2008.

\bibitem{dodge2018quality}
Samuel Dodge and Lina Karam.
\newblock Quality robust mixtures of deep neural networks.
\newblock {\em IEEE Transactions on Image Processing}, 27(11):5553--5562, 2018.

\bibitem{gopakumar2018algorithmic}
Shivapratap Gopakumar, Sunil Gupta, Santu Rana, Vu Nguyen, and Svetha
  Venkatesh.
\newblock Algorithmic assurance: An active approach to algorithmic testing
  using bayesian optimisation.
\newblock In {\em NIPS}, pages 5466--5474, 2018.

\bibitem{gotovos2013lse}
Alkis Gotovos, Nathalie Casati, Gregory Hitz, and Andreas Krause.
\newblock Active learning for level set estimation.
\newblock In {\em IJCAI}, pages 1344--1350, 2013.

\bibitem{ha2021high}
Huong Ha, Sunil Gupta, Santu Rana, and Svetha Venkatesh.
\newblock {High Dimensional Level Set Estimation with Bayesian Neural Network}.
\newblock In {\em AAAI}, volume~35, pages 12095--12103, 2021.

\bibitem{he2016deep}
Kaiming He, Xiangyu Zhang, Shaoqing Ren, and Jian Sun.
\newblock Deep residual learning for image recognition.
\newblock In {\em CVPR}, pages 770--778, 2016.

\bibitem{heusel2017gans}
Martin Heusel, Hubert Ramsauer, Thomas Unterthiner, Bernhard Nessler, and Sepp
  Hochreiter.
\newblock Gans trained by a two time-scale update rule converge to a local nash
  equilibrium.
\newblock {\em NIPS}, 30, 2017.

\bibitem{zhang2018mixup}
Zhang Hongyi, Cisse Moustapha, Dauphin Yann, and Lopez-Paz David.
\newblock mixup: Beyond empirical risk minimization.
\newblock In {\em ICLR}, 2018.

\bibitem{hossain2019distortion}
Md~Tahmid Hossain, Shyh~Wei Teng, Dengsheng Zhang, Suryani Lim, and Guojun Lu.
\newblock Distortion robust image classification using deep convolutional
  neural network with discrete cosine transform.
\newblock In {\em IEEE International Conference on Image Processing (ICIP)},
  pages 659--663, 2019.

\bibitem{inatsu2019active}
Yu Inatsu, Masayuki Karasuyama, Keiichi Inoue, and Ichiro Takeuchi.
\newblock Active learning for level set estimation under cost-dependent input
  uncertainty.
\newblock {\em arXiv preprint arXiv:1909.06064}, 2019.

\bibitem{inatsu2020active}
Yu Inatsu, Masayuki Karasuyama, Keiichi Inoue, and Ichiro Takeuchi.
\newblock Active learning for level set estimation under input uncertainty and
  its extensions.
\newblock {\em Neural Computation}, 32(12):2486--2531, 2020.

\bibitem{ioffe2015batch}
Sergey Ioffe and Christian Szegedy.
\newblock Batch normalization: Accelerating deep network training by reducing
  internal covariate shift.
\newblock In {\em ICML}, pages 448--456, 2015.

\bibitem{kang2014convolutional}
Le Kang, Peng Ye, Yi Li, and David Doermann.
\newblock Convolutional neural networks for no-reference image quality
  assessment.
\newblock In {\em CVPR}, pages 1733--1740, 2014.

\bibitem{leevy2018survey}
Joffrey Leevy, Taghi Khoshgoftaar, Richard~A Bauder, and Naeem Seliya.
\newblock A survey on addressing high-class imbalance in big data.
\newblock {\em Journal of Big Data}, 5(1):1--30, 2018.

\bibitem{li2019blind}
Xiaoyu Li, Bo Zhang, Pedro Sander, and Jing Liao.
\newblock Blind geometric distortion correction on images through deep
  learning.
\newblock In {\em CVPR}, pages 4855--4864, 2019.

\bibitem{mirza2014conditional}
Mehdi Mirza and Simon Osindero.
\newblock Conditional generative adversarial nets.
\newblock {\em arXiv preprint arXiv:1411.1784}, 2014.

\bibitem{nguyen2022fskd}
Dang Nguyen, Sunil Gupta, Kien Do, and Svetha Venkatesh.
\newblock Black-box few-shot knowledge distillation.
\newblock In {\em ECCV}, 2022.

\bibitem{nguyen2021knowledge}
Dang Nguyen, Sunil Gupta, Trong Nguyen, Santu Rana, Phuoc Nguyen, Truyen Tran,
  Ky Le, Shannon Ryan, and Svetha Venkatesh.
\newblock Knowledge distillation with distribution mismatch.
\newblock In {\em ECML-PKDD}, pages 250--265, 2021.

\bibitem{rawat2017deep}
Waseem Rawat and Zenghui Wang.
\newblock Deep convolutional neural networks for image classification: A
  comprehensive review.
\newblock {\em Neural Computation}, 29(9):2352--2449, 2017.

\bibitem{sampath2021survey}
Vignesh Sampath, I{\~n}aki Maurtua, Juan~Jose Aguilar~Martin, and Aitor
  Gutierrez.
\newblock A survey on generative adversarial networks for imbalance problems in
  computer vision tasks.
\newblock {\em Journal of Big Data}, 8:1--59, 2021.

\bibitem{snoek2012practical}
Jasper Snoek, Hugo Larochelle, and Ryan Adams.
\newblock {Practical Bayesian optimization of machine learning algorithms}.
\newblock In {\em NIPS}, pages 2951--2959, 2012.

\bibitem{sohn2015learning}
Kihyuk Sohn, Honglak Lee, and Xinchen Yan.
\newblock Learning structured output representation using deep conditional
  generative models.
\newblock {\em NIPS}, pages 3483--3491, 2015.

\bibitem{thai2010cost}
Nguyen Thai-Nghe, Zeno Gantner, and Lars Schmidt-Thieme.
\newblock Cost-sensitive learning methods for imbalanced data.
\newblock In {\em IJCNN}, pages 1--8, 2010.

\bibitem{tian2020contrastive}
Yonglong Tian, Dilip Krishnan, and Phillip Isola.
\newblock Contrastive representation distillation.
\newblock In {\em ICLR}, 2020.

\bibitem{van2008visualizing}
Laurens Van~der Maaten and Geoffrey Hinton.
\newblock {Visualizing data using t-SNE}.
\newblock {\em Journal of Machine Learning Research}, 9(11):2579--2605, 2008.

\bibitem{wang2020neural}
Dongdong Wang, Yandong Li, Liqiang Wang, and Boqing Gong.
\newblock {Neural Networks Are More Productive Teachers Than Human Raters:
  Active Mixup for Data-Efficient Knowledge Distillation from a Blackbox
  Model}.
\newblock In {\em CVPR}, pages 1498--1507, 2020.

\bibitem{wang2020generalizing}
Yaqing Wang, Quanming Yao, James Kwok, and Lionel Ni.
\newblock Generalizing from a few examples: A survey on few-shot learning.
\newblock {\em ACM Computing Surveys}, 53(3):1--34, 2020.

\bibitem{xu2022generating}
Jingyi Xu and Hieu Le.
\newblock Generating representative samples for few-shot classification.
\newblock In {\em CVPR}, pages 9003--9013, 2022.

\bibitem{yin2020dreaming}
Hongxu Yin, Pavlo Molchanov, Jose Alvarez, Zhizhong Li, Arun Mallya, Derek
  Hoiem, Niraj Jha, and Jan Kautz.
\newblock {Dreaming to distill: Data-free knowledge transfer via
  DeepInversion}.
\newblock In {\em CVPR}, pages 8715--8724, 2020.

\bibitem{zhou2017classification}
Yiren Zhou, Sibo Song, and Ngai-Man Cheung.
\newblock On classification of distorted images with deep convolutional neural
  networks.
\newblock In {\em IEEE International Conference on Acoustics, Speech and Signal
  Processing (ICASSP)}, pages 1213--1217, 2017.

\end{thebibliography}
